\documentclass{article} 
\usepackage{iclr2026_conference,times}

\usepackage{amsmath,amssymb,amsfonts}
\usepackage{algorithmic}
\usepackage{graphicx}
\usepackage{textcomp}
\usepackage[title]{appendix}%
\usepackage{xcolor}
\usepackage{xurl}
\usepackage{booktabs}
\usepackage{multirow}
\usepackage{array}
\usepackage{listings}
\usepackage{booktabs}
\usepackage{caption}
\usepackage{afterpage}
\usepackage{tabularx}
\usepackage{cuted}
\usepackage{comment}

\usepackage{appendix}

\DeclareCaptionType{prompt}[\textbf{Prompt}][List of Prompts]

\usepackage[colorlinks]{hyperref}
\usepackage{subcaption}
\hypersetup{
    linkcolor=blue,
    filecolor=magenta,
    urlcolor=teal,
    citecolor=magenta
    }

\title{\raggedright{ECGQuest: Benchmarking and Fine-Tuning Language Models for Electrocardiography}}
\author{
Mohammadsina Hassannia \\
Department of Biomedical Informatics \\
Emory University \\
Atlanta, GA 30322, USA \\
\url{sina.hassannia@dbmi.emory.edu}
\And
Matthew A. Reyna \\
Department of Biomedical Informatics \\
Emory University \\
Atlanta, GA 30322, USA \\
\url{matthew@dbmi.emory.edu}
\And
Reza Sameni\\
Department of Biomedical Engineering \\
Georgia Institute of Technology \\
and Department of Biomedical Informatics\\
Emory University \\
Atlanta, GA 30332, USA \\
\url{rsameni@emory.edu}
}

\iclrfinalcopy
\begin{document}

\maketitle

\begin{abstract}

Electrocardiogram (ECG) interpretation requires knowledge of cardiology, electrophysiology, clinical diagnosis, ECG waveforms, signal acquisition, and ECG instrumentation. While large language models have proven effective for such applications, existing language-models  primarily assess broad medical knowledge or the interpretation of individual ECG signals and images, rather than comprehensive context knowledge required for ECG interpretation. We built ECGQuest, a literature-grounded resource for evaluating and fine-tuning ECG-specific language models. An automated GPT-4o-based pipeline generated questions from 23 ECG references and proceedings of the Computing in Cardiology conference between 2003--2025. The final dataset contains 10{,}904 unique True/False questions paired with their negated form derived from the same source (21{,}808 Q\&A in total). We evaluated three commercial and 20 open-source language models on the held-out ECGQuest test set in a zero-shot setting. We also fine-tuned five open-source models with 7--14\,B parameters using Low-Rank Adaptation and included BERT and BiomedBERT as supervised encoder baselines. Generalization was evaluated using ECG-related subsets of MedMCQA and MedQA datasets converted to binary True/False questions using their official examination answer keys. Zero-shot accuracy on ECGQuest ranged from 49.5\% to 74.4\%, with GPT-5 achieving the highest accuracy. General-purpose models outperformed all medically specialized models, and several models showed strong directional True/False biases. Encoder baselines performed near chance. Fine-tuning improved all open-source models on ECGQuest by 6.5--14.1\%. The fine-tuned DeepSeek-R1-Distill-Qwen-14\,B model achieved 76.3\% accuracy, and a five-model voting ensemble achieved the highest overall accuracy of 78.5\%. On MedMCQA and MedQA, fine-tuning primarily benefited weak or class-biased models and did not consistently improve already strong base models. ECGQuest is a reproducible benchmark for evaluating contextual ECG knowledge in language models and demonstrates that targeted, parameter-efficient fine-tuning can make small language models competitive with substantially larger commercial models. 

\end{abstract}

\section{Introduction}    

The electrocardiogram (ECG) is among the most widely used diagnostic tools in cardiovascular medicine, routinely used across emergency, inpatient, and outpatient settings to detect arrhythmias, ischemia, conduction abnormalities, and a wide range of other cardiac conditions. Despite the broad use of the ECG, its accurate interpretation is challenging and requires substantial training and experience in cardiac electrophysiology. Moreover, even experienced ECG annotators show significant inter-rater and intra-rater variability, leading to missed diagnoses, inappropriate or unnecessary interventions, and delayed treatments~\citep{cook2020ecg,rafie2021ecg}. The combination of high clinical volume and persistent interpretive challenges has motivated the development of computational systems capable of reliable and consistent ECG interpretation.

ECG interpretation is a multifaceted process rather than a binary or multiclass classification skill. It requires basic knowledge of the electrophysiology of the heart, ECG acquisition technologies, analog and digital frontend requirements, sampling and filtering specifications. It also involves quantitative measurements derived from the waveform, such as heart rate, rhythm, and interval durations including the PR, QRS, and QT intervals. Moreover, ECG diagnosis is descriptive and multi-labeled, where patterns are identified, associated with their electrophysiological mechanisms, and integrated with the clinical and personalized context to form a conclusion. The measurement component is primarily programmatic and conducted automatically via software, whereas the descriptive component relies on holistic knowledge of electrophysiology, cardiology and the subject's background.

Overall, the knowledge base required for accurate ECG diagnosis is broad and cross-disciplinary, motivating the use of methods that can integrate information across domains. LLMs have been promising in similar applications. They encode broad knowledge spanning natural sciences to engineering~\citep{brown2020gpt3,hendrycks2021mmlu}, and have reached close to expert-level performance on similar clinical tasks (cf. Section~\ref{sec:related}). Despite this progress, current evaluations of clinical language models focus either on broad medical knowledge or on the interpretation of specific ECG anomalies. As a result, the background and contextual knowledge of the ECG, beyond specific ECG anomalies, is often not considered by existing LLMs. Therefore, it remains unclear how much a language model actually \textit{``understands''} an ECG, rather than only classifying or describing patterns.

To address this gap, we made the following contributions: (i) We used commercial LLMs to parse ECG-related literature and constructed a text-only database of general ECG questions and answers comprising $10{,}904$ True/False question pairs ($21{,}808$ questions in total). The questions were derived from $23$ scientific ECG resources and specialized ECG conference proceedings, spanning various ECG-related topics such as clinical, electro-physiological, biomedical signal processing, ECG diagnosis, and ECG hardware and electronics. (ii) We evaluated a diverse set of proprietary, general-purpose, biomedical, and open-access language models on this benchmark in a zero-shot setting. (iii) We explored whether smaller models in the $7$ to $14$ billion parameter range can be specialized for ECG knowledge using low-rank adaptation. (iv) We extended the evaluation to external ECG-related test sets extracted from MedQA (USMLE style)~\citep{jin2021medqa} and MedMCQA~\citep{pal2022medmcqa}, whose labels come from official examination keys and are therefore independent of our generation pipeline; this analysis is reported in Supplement~\ref{app:external}. 

\section{Existing Medical Q\&A and Domain-Specific Language Models} 
\label{sec:related}
Multiple-choice benchmarks, such as MedQA~\citep{jin2021medqa}, MedMCQA~\citep{pal2022medmcqa}, and PubMedQA \citep{jin2019pubmedqa} have been used as references to evaluate medical language models. On these benchmarks, both general-purpose and domain-adapted models, including Med-PaLM~\citep{singhal2023clinical}, Clinical Camel~\citep{toma2023clinicalcamel}, and BioInstruct~\citep{tran2024bioinstruct}, have demonstrated strong, and, in some cases expert-level, accuracy through prompting, instruction tuning, and parameter efficient fine-tuning. However, these benchmarks assess broad clinical knowledge rather than ECG knowledge; Health-LLM, for instance, focuses on physiological time-series from wearable sensors rather than the conceptual knowledge required for ECG interpretation. Accuracy on these benchmarks therefore provides limited insight into a model's understanding of ECG.

Recent research have increasingly integrates language models in ECG applications. ECG-QA~\citep{oh2023ecgqa} paired the PTB-XL and MIMIC-IV ECG datasets, with seventy expert-validated question templates spanning verify, choose, and query formats, and evaluates language models by translating signal model outputs into text. Q-HEART~\citep{pham2025qheart}, addressed ECG-related question answering through a knowledge-informed multimodal LLM that combines ECG representations, retrieved clinical context, and instruction tuning, achieving state-of-the-art performance on ECG-QA. Using ECG images, ECGInstruct~\citep{liu2024pulse} compiled over one million ECG image instruction-tuning samples and trains PULSE, an open multimodal model for ECG image interpretation. PULSE is evaluated on the ECGBench dataset across feature recognition, rhythm analysis, morphology assessment, and report generation. MEIT~\citep{wan2025meit} instruction-tuned nine language models for ECG report generation by integrating signal representations into the model's latent space and benchmarked them on more than 800,000 reports. 

\section{Methods}
\begin{figure*}[t]
  \centering
  \includegraphics[width=\textwidth]{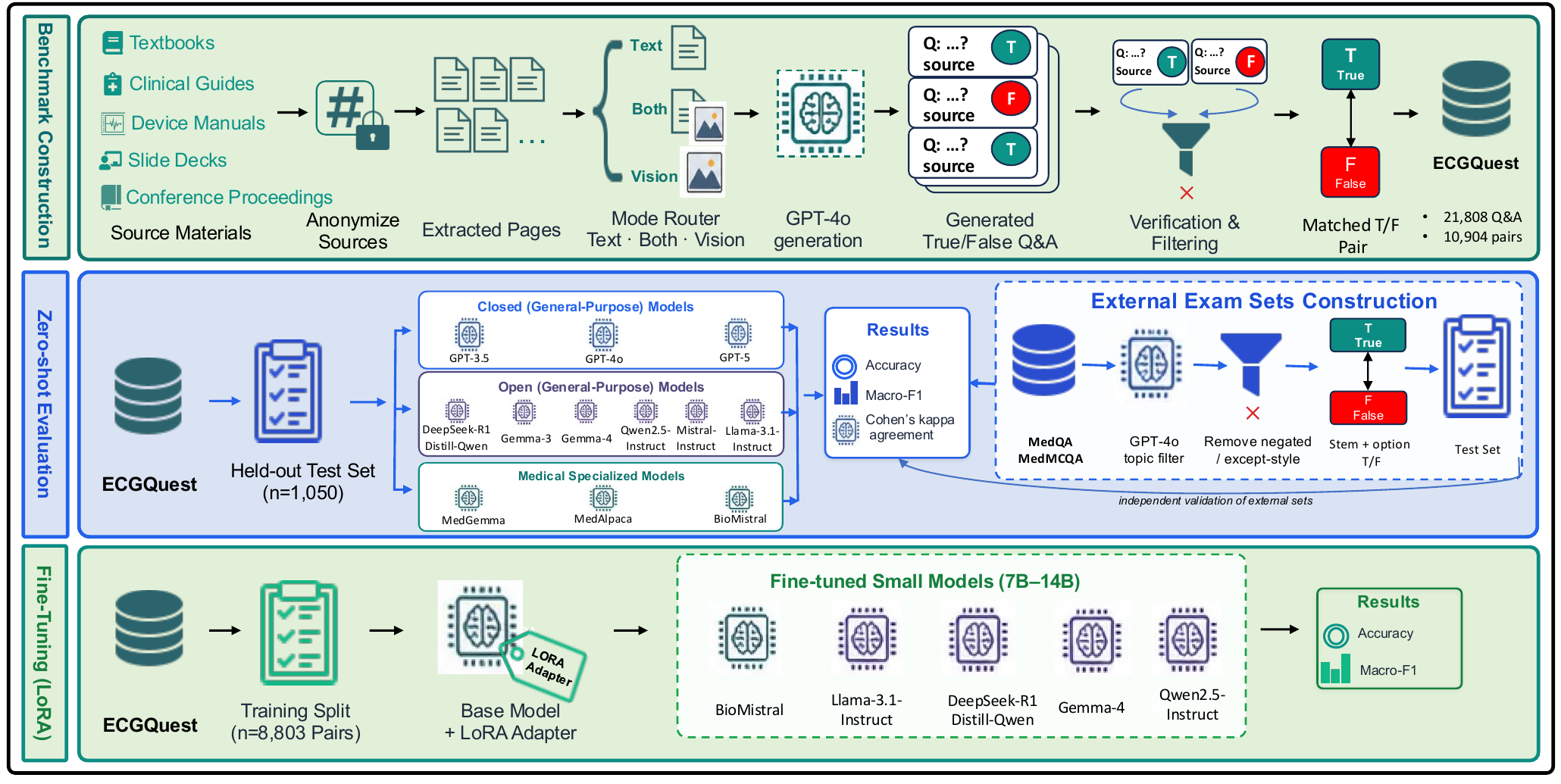}
  \caption{\small{Overview of ECGQuest creation framework. \textit{(1)} ECG reference materials were anonymized, split into pages, routed by content to a text or vision-based generation model for text extraction, and passed to GPT-4o to generate paired True/False questions. \textit{(2)} Closed, open, and medically specialized models were tested on the held-out test set ($n{=}1{,}050$), and external ECG-focused MedQA and MedMCQA questions, extracted by GPT-4o ECG-topic filtering and converted to True/False for independent validation. \textit{(3)} small open models (7--14\,B parameters) were fine-tuned with LoRA adapters and re-evaluated.}}\vspace{-0.1in}
  \label{fig:overview}
\end{figure*}

Figure~\ref{fig:overview} summarizes the three stages of our study: (i) Constructing ECGQuest, a literature-grounded benchmark question and answer (Q\&A) resource from the ECG literature; (ii) evaluating a broad set of closed source, open source, and medically specialized language models in a zero-shot setting on a held-out test set, and, as an independent check of generalization, on two external exam-labeled ECG question sets whose construction and results are given in Supplement~\ref{app:external}; (iii) fine-tuning compact open source models using Low-Rank Adaptation (LoRA)~\citep{hu2021lora}.

\subsection{ECGQuest Dataset Construction}
\label{sec:data_cons}
An ECG-specific question-answer benchmark was constructed by converting a curated corpus of electrocardiography reference materials into balanced, labeled True/False items. The construction pipeline consists of four stages: (i) corpus assembly and text extraction, (ii) question generation, (iii) label balancing through paired negation, (iv) and validation and deduplication. Each stage is fully automated, and each generated item is linked to its source, ensuring that question correctness depends on the reference text itself, and not on the prior knowledge of the model.

\textit{Data Source:} The input corpus used for Q\&A extraction comprises $23$ electrocardiography reference documents from five source categories: (1) textbooks \citep{hall2020guyton,gertsch2004ecg,kamath_hrv,hampton2019ecg, thaler2023ekg,jevon2009ecgs,brunner_suddarth_handbook}, (2) clinical guidelines and manuals \citep{vera2024ekg, rawshani_ecg, lifesaver_ecg_rhythm, philips_dxl, ge12sl}, (3) ECG device manuals \citep{philips2023tc10,philips2019tc50tc70,ge2016musev9,nihonkohden_ecg1350,cardiacinsight2025cardea,welchallyn2023rscribe,schiller1992at6,schiller2020at102g2,jhs2009ecg}, (4) online clinical tutorials and teaching materials \citep{yanowitz_ecg,patrick_ecg_step}, and (5) full-text conference proceedings from the Computing in Cardiology (CinC) conference published between 2003 and 2025
\citep{CinC2003,CinC2004,CinC2005,CinC2006,CinC2007,CinC2008, CinC2009,CinC2010,CinC2011,CinC2012,CinC2013,CinC2014, CinC2015,CinC2016,CinC2017,CinC2018,CinC2019,CinC2020, CinC2021,CinC2022,CinC2023,CinC2024,CinC2025}. These resources span the full range of ECG literature, from foundational cardiac electrophysiology and clinical interpretation texts to manufacturer documentation. Together, they cover ECG knowledge at multiple levels, including waveform morphology, clinical diagnosis, electrophysiological mechanisms, acquisition procedures, instrumentation, and device-specific technical specifications.

\textit{Data Preprocessing:}
\label{sec:page_ext}
All source files were in vector PDF format. In order to keep track of the integrity of the dataset, each source PDF was renamed to its 128-bit MD5 (Message-Digest Algorithm 5; RFC 1321) checksum before processing, computed over the raw file bytes and rendered as a 32-character hexadecimal string. Because all intermediate stages operated on hashed identifiers, the generation pipeline was blind to the source title. Each hashed document was programmatically split into separate single-page PDF files. From each page, we extracted the text using \texttt{pdfplumber} \citep{SingerVine2015}. The extracted text was then serialized into XML files that include source reference MD5 checksum, reference category, page number, character count, and text content. The character count was calculated at this stage and subsequently used as an indicator to decide how each page should be processed for question generation (due to LLM and SLM token size limitations). Pages without extractable text, typically image-only pages such as full-page ECG tracings, were logged separately and directed to the vision-based generation path described below. 

\textit{Q\&A Generation:}
Questions were generated from the extracted text using a GPT-4o, which operates in one of three modes determined automatically by the character count of each page extracted in Section ~\ref{sec:page_ext}. Pages containing 400 or more characters were processed in text mode, using only the extracted page text. Pages with 100 to 399 characters were processed in both modes, utilizing both the extracted text and a rendered image of the page, as sparse pages frequently contain essential information in figures, tables, or captions. Pages with fewer than 100 characters were processed in vision mode, utilizing only the rendered page image, which is appropriate for image-dominant pages with minimal extractable text. In all three modes, the model generated factual statements limited to four ECG topics: (i) waveform features (e.g., P wave, QRS complex, ST segment, T and U waves, intervals, morphology, etc.); (ii) clinical interpretation and diagnostic significance of ECG patterns (e.g., ischemia, infarction, arrhythmia, conduction blocks, hypertrophy, electrolyte effects, and normal versus abnormal decision criteria, etc.); (iii) the physiological basis of ECG findings (e.g., conduction system, depolarization and repolarization, action potentials, automaticity, refractory periods, etc.); and (iv) ECG device specifications (e.g., leads, electrodes, gain, filters, paper speed, calibration, artifacts, etc.). Each generated item was combined into a structured form including the page number, the question, a True/False label, and the original source text.

As shown Supplement~\ref{app:prompts}, GPT-4o was instructed to deliver questions from the source text alone, without reliance on external medical knowledge or requiring viewing an image (for example, ``the rhythm shown, below'', or similar that appears in many references and manuals). This constraint was to ensure that every question can be answered using clinical knowledge available in the provided text alone, independent of an image. Since the instructions in the prompt are not necessarily followed by the language model, the generated items were passed through a postprocessing verification stage (again conducted by a language model) to evaluate each item against these criteria: an item was rejected if its source text could not be found on the source page; if its question referenced the source figure as detected through a dictionary of image-referencing terms such as \emph{shown}, \emph{figure}, or \emph{this tracing}; or if it was malformed, missing a required field, or contained an ambiguous label other than True or False. In total, 638 prompt responses were rejected at this stage. Representative examples and the reasons for their removal are provided in the Supplement~\ref{app:prompts}. For pages processed in both modes, text- and image-conditioned outputs were de-duplicated by another batch post-processing to remove redundant items. Each remaining item retains its origin, including source document MD5, page number, and generation mode, ensuring that every question can be traced to its originating page. The prompts used to generate the dataset are provided in Supplement~\ref{app:prompts}.

\textit{True-False Label Balancing:} 
\label{sec:tf_balance}
True-False questions that passed the verification stages in Section~\ref{sec:data_cons} had an imbalanced label distributions, as individual pages can contain true or false statements depending on their content. To balance the label distribution, each validated question was paired with its negative form derived from the same source text, to ensure that the dataset contains an equal number of true and false questions. For each true question, GPT-4o was prompted to generate a false counterpart that contradicts the source text by altering a minor clinical detail or by paraphrasing while preserving the source text (by negating an assertion or substituting an incorrect value, direction, or anatomical or temporal term). Conversely, for each false question, the model was prompted to produce the corresponding true statement by restating the fact as expressed in the source text. In both cases, the counterpart was derived solely from the source text, without introducing any concepts not present in the original; so that the pair differs primarily in the detail determining the label. The complete label-balancing prompt is provided in Supplement~\ref{app:prompts}.

To verify the negation process, each counterpart question was verified programmatically: (i) The source text must remain identical to the original item; (ii) the counterpart must have the correctly flipped label, differ from the original. Counterparts failing any of these criteria were discarded. This process resulted in question-answer pairs with balanced label distribution, where every true/false statement is paired with a minimally edited false/true version of the same question. This pairing prevents models from exploiting irrelevant clues, such as true/false prevalence or topic, as shortcuts to the label. The resulting pairs were de-duplicated in a final stage to remove near-identical facts repeated across documents. Two pairs were considered duplicates if the Jaccard similarity of their normalized source text (in lowercase and stripped of punctuation and whitespace) was at least 0.90. Because the sources overlap in content, this de-duplication step, prevented redundant questions across training/testing data splits.

The final ECGQuest corpus includes $10{,}904$ unique question pairs (or $21{,}808$ individual True/False questions) with a balanced $50\%$ True-False label distribution. GPT-4o was further used to categorize the questions under four categories. As shown in Table~\ref{tab:dataset_stats} in the Supplement, $6735$ pairs are related to ECG diagnosis, $1917$ pairs to ECG waveform characteristics, $1200$ pairs to ECG acquisition technologies, and $1052$ pairs to basic cardiac physiology. 

\subsection{Model Fine-Tuning}
For fine-tuning, the dataset was partitioned into training, validation, and test splits in an approximately $80{:}10{:}10$ ratio, with both members of each True/False pair assigned to the same split. All models were fine-tuned targeting maximum macro-F1 score, which due to the balanced True/False set of ECGQuest questions is technically identical to maximizing accuracy. The two categories of fine-tuned models are detailed below.

\textit{Encoder-based model fine-tuning:} As non-generative baselines, two pre-trained encoders were fine-tuned as binary True/False classifiers: BERT~\citep{devlin2019bert} and BiomedBERT~\citep{gu2021domain}. Each model was fine-tuned with a two-class classification on the same training split as all other models and evaluated on the identical test sets. In contrast to the generative models, no prompt, chat template, or text generation was involved. These classifiers establish a lower bound, indicating how much of the task can be solved by direct supervised classification.


\textit{Instruction-based model fine-tuning:} BioMistral-7\,B, Llama-3.1-8\,B, Qwen2.5-14\,B, Gemma-4-12\,B, and DeepSeek-R1-14\,B were fine-tuned on the training split ($8{,}803$ question pairs) and evaluated on the held-out test split ($n=1{,}050$). All models were adapted using LoRA~\citep{hu2021lora} on the attention and MLP projections, with an effective batch size of 64, a dropout of 0.1, and LoRA $\alpha=2r$. BioMistral-7\,B used a LoRA rank of 16 and a False-class weight of 1.5, whereas Llama-3.1-8\,B, Qwen2.5-14\,B, Gemma-4-12\,B, and DeepSeek-R1-14\,B used a rank of 64 and False-class weights of 3.0, 0.8, 1.0, and 1.5, respectively. The class-weighted loss was used to reduce each base model's tendency to overpredict either True or False.

\section{Results}

\subsection{Zero-shot Benchmarking of LLMs and SLMs on ECGQuest}   
We evaluated a comprehensive set of language models on the held-out test split (n=1{,}050) in a zero-shot setting, with no weight updates, to measure how well existing models answer ECG True/False questions. The models span three groups: closed-source general-purpose models, open-source general-purpose models, and medically specialized open-source models. Each question was presented with a short instruction and requests a single True or False answer. The models, their parameter counts, and performance across multiple metrics are presented in the upper part of Table~\ref{tab:zeroshot}.

Zero-shot accuracy ranges from 49.5\% to 74.4\%, with no model exceeding 75\%. GPT-5 performs best (74.4\% accuracy; 74.1\% macro-F1), followed by GPT-4o (71.7\%; 71.7\%) and the leading open-source model, Gemma-4-31\,B (71.0\%; 70.9\%). The gap between the best closed-source and best open-source model was 3.4\% accuracy points. Gemma-4-31\,B is within 0.7 points of GPT-4o at a small fraction of its estimated parameter count.

The gap between accuracy and macro-F1 identifies directional response biases. MedAlpaca-7\,B (sensitivity 93.4, specificity 6.0; macro-F1 38.0 versus accuracy 50.2\%) and BioMistral-7\,B (sensitivity 90.2, specificity 18.7; macro-F1 48.0 versus accuracy 54.9\%) are biased toward True, while Gemma-3-270\,M (sensitivity 0.4, specificity 99.8; macro-F1 33.5 versus accuracy 49.5\%) and Qwen2.5-Instruct-14\,B (sensitivity 27.5, specificity 93.6; macro-F1 55.5 versus accuracy 60.2\%) are biased toward False. In contrast, GPT-5, GPT-4o, Llama-3.1-Instruct-8\,B, MedGemma-4\,B, DeepSeek-R1-Distill-Qwen-14\,B, and the larger Gemma-3 and Gemma-4 models have accuracy and macro-F1 within 0.3 points of each other, indicating that their accuracy reflects discrimination between correct and incorrect ECG statements rather than a directional answering bias.

Between models of the same family, with varied number of parameters, accuracy increases consistently. Gemma-3 improves monotonically from 49.5\% at 270\,M, 50.2\% at 1\,B, 56.4\% at 4\,B, 64.6\% at 12\,B, to 67.1\% at 27\,B; Gemma-4 is non-decreasing over the same range (58.1\% at 2\,B, 62.9\% at 4\,B and 12\,B, 66.5\% at 26\,B, 71.0\% at
31\,B), with a plateau between 4\,B and 12\,B; DeepSeek-R1-Distill-Qwen shows the same pattern (53.2\% at 1.5\,B, 59.0\% at 7\,B, 69.8\% at 14\,B). The same pattern is observed between closed-source models.

Interestingly, medical specialization of the models did not necessarily enhance ECG knowledge. The performance of medical models ranges from $50.2\%$ (MedAlpaca-7\,B) to $59.5\%$ (MedGemma-4\,B), with eight open general purpose models outperforming the best medical model. MedGemma-4\,B is the only medical model that achieves the performance of mid sized general models, yet it remains below Gemma-4-4\,B (62.9\%) at a comparable size.

\subsection{Fine-tuned Model Performances}
\textit{Encoder-based models:} As non-generative baselines, the fine-tuned versions of BERT~\citep{devlin2019bert} and BiomedBERT~\citep{gu2021domain} both perform near the 50\% chance level. BERT achieves 52.1\% and BiomedBERT achieves 54.0\%. These near-chance results highlight the poor performance of these models for ECG-related tasks.

\textit{Fine-tuned small language models:} The lower part of Table~\ref{tab:zeroshot} presents results for the five fine-tuned language models and a voting-based ensemble built from them. Fine-tuning outperformed all models on ECGQuest, by between 6.5 and 14.1 percentage points, enabling $7$--$14$\,B models match or exceed GPT-5 and GPT-4o, the two strongest models in the zero-shot setting, despite substantially fewer parameters. DeepSeek-R1-14\,B reaches 76.3\% (macro-F1 76.1), outperforming GPT-5 (74.4\%), while Qwen2.5-14\,B (74.0\%) and Gemma-4-12\,B ($73.3\%$) both surpass GPT-4o. Llama-3.1-8\,B ($71.7\%$) matches GPT-4o, and BioMistral-7\,B, the weakest base in the group at 54.9\%, gains 14.1 points to reach $69.0\%$, the largest improvement among the fine-tuned models. We also evaluated an ensemble model by majority voting between the five adapted models. This resulted in the best result on ECGQuest: 78.5\% accuracy with macro-F1 78.4. All five are balanced, with closely aligned accuracy and macro-F1, and all outperform the encoder baselines reported in the previous section.

\textit{Generalization to external exam-labeled questions:} To test whether these gains reflect transferable ECG knowledge rather than adaptation  ECGQuest, we evaluated every model on ECG-related subsets of MedMCQA~\citep{pal2022medmcqa} and MedQA~\citep{jin2021medqa}, converted to balanced True/False items using their official examination keys. The construction procedure is described in Supplement~\ref{app:external-prep} and the complete
results are reported in Table~\ref{tab:external_table} of Supplement~\ref{app:external}. The zero-shot ranking is broadly preserved across all three sets, but the effect of fine-tuning depends strongly on the base model: it improves weak or
class-biased models, such as BioMistral-7\,B (51.3\% to 60.5\% on MedMCQA) and Qwen2.5-Instruct-14\,B (67.3\% to 73.0\%), while reducing accuracy for the already balanced Llama-3.1-Instruct-8\,B and DeepSeek-R1-14\,B. We discuss this asymmetry in Section~\ref{sec:discussion}.

\begin{table*}[tb]
\centering
\caption{Zero-shot and fine-tuned language model performances on ECGQuest ($n{=}1{,}050$). Acc: accuracy; F1: macro-averaged F1 score;
P: precision; R/Sen: recall/sensitivity; Spec: specificity.
Across all models, the best value per metric column is in \textbf{bold} and the
runner-up is \underline{underlined}.}
\label{tab:zeroshot}

\begin{tabular}{clrccccc}
\toprule
 & & & \multicolumn{5}{c}{\textbf{ECGQuest}} \\
\cmidrule(lr){4-8}

 & \textbf{Model} 
 & \textbf{Params} 
 & \textbf{Acc (\%)} 
 & \textbf{F1} 
 & \textbf{P} 
 & \textbf{R/Sen} 
 & \textbf{Spec} \\
\midrule

\multirow{22}{*}{\rotatebox[origin=c]{90}{\textbf{Zero-shot learning}}}
 & \multicolumn{7}{l}{\textit{General purpose (closed source)}} \\

 & GPT-5    & 2\,T\textsuperscript{*}   
 & 74.4 & 74.1 & \textbf{81.8} & 63.5 & 85.5 \\

 & GPT-4o   & 200\,B\textsuperscript{*} 
 & 71.7 & 71.7 & 72.8 & 70.4 & 73.0 \\

 & GPT-3.5  & 175\,B\textsuperscript{*} 
 & 63.1 & 61.7 & 60.0 & 81.4 & 44.5 \\

\cmidrule(l){2-8}

 & \multicolumn{7}{l}{\textit{General-purpose (open source)}} \\

 & Mistral-Instruct-v0.3 & 7\,B
 & 57.6 & 57.4 & 57.2 & 63.6 & 51.4 \\

 & Llama-3.1-Instruct & 8\,B
 & 58.4 & 58.4 & 59.3 & 56.3 & 60.5 \\

 & Qwen2.5-Instruct & 14\,B
 & 60.2 & 55.5 & \underline{81.6} & 27.5 & \underline{93.6} \\

 & DeepSeek-R1-Distill-Qwen & 1.5\,B
 & 53.2 & 52.9 & 53.6 & 59.1 & 47.0 \\

 & DeepSeek-R1-Distill-Qwen & 7\,B
 & 59.0 & 58.1 & 60.5 & 68.1 & 48.4 \\

 & DeepSeek-R1-Distill-Qwen & 14\,B
 & 69.8 & 69.8 & 71.4 & 70.5 & 69.1 \\

 & Gemma-3 & 270\,M
 & 49.5 & 33.5 & 66.7 & 0.4 & \textbf{99.8} \\

 & Gemma-3 & 1\,B
 & 50.2 & 50.0 & 50.9 & 43.3 & 57.2 \\

 & Gemma-3 & 4\,B
 & 56.4 & 53.2 & 54.6 & 81.4 & 30.8 \\

 & Gemma-3 & 12\,B
 & 64.6 & 64.4 & 63.3 & 71.4 & 57.6 \\

 & Gemma-3 & 27\,B
 & 67.1 & 67.1 & 69.2 & 63.1 & 71.3 \\

 & Gemma-4 & 2\,B
 & 58.1 & 56.1 & 65.4 & 36.3 & 80.3 \\

 & Gemma-4 & 4\,B
 & 62.9 & 62.7 & 66.0 & 54.8 & 71.1 \\

 & Gemma-4 & 12\,B
 & 62.9 & 62.7 & 62.9 & 64.4 & 61.1 \\

 & Gemma-4 & 26\,B
 & 66.5 & 66.4 & 68.4 & 62.7 & 70.3 \\

 & Gemma-4 & 31\,B
 & 71.0 & 70.9 & 72.8 & 68.0 & 74.0 \\

\cmidrule(l){2-8}

 & \multicolumn{7}{l}{\textit{Medical-specialized (open source)}} \\

 & MedGemma & 4\,B
 & 59.5 & 59.5 & 59.8 & 61.0 & 58.0 \\

 & MedAlpaca & 7\,B
 & 50.2 & 38.0 & 50.4 & \textbf{93.4} & 6.0 \\

 & BioMistral & 7\,B
 & 54.9 & 48.0 & 53.2 & \underline{90.2} & 18.7 \\

 & MedAlpaca & 13\,B
 & 55.1 & 53.5 & 54.2 & 73.3 & 36.6 \\

\midrule

 & \multicolumn{7}{l}{\textit{Encoder baselines}} \\

 & BERT & 110\,M
 & 52.1 & 52.1 & 52.7 & 52.4 & 51.8 \\

 & BiomedBERT & 110\,M
 & 54.0 & 53.5 & 55.8 & 43.5 & 64.7 \\

\cmidrule(l){2-8}

\multirow{8}{*}{\rotatebox[origin=c]{90}{\textbf{Fine-tuned}}}
 & \multicolumn{7}{l}{\textit{Decoder LLMs}} \\

 & BioMistral & 7\,B
 & 69.0 & 68.9 & 68.7 & 70.8 & 67.1 \\

 & Llama-3.1-Instruct & 8\,B
 & 71.7 & 71.7 & 71.7 & 71.1 & 71.7 \\

 & Gemma-4 & 12\,B
 & 73.3 & 73.0 & 69.6 & 84.0 & 62.4 \\

 & DeepSeek-R1-Distill-Qwen & 14\,B
 & \underline{76.3} & \underline{76.1} & 74.9 & 82.2 & 69.9 \\

 & Qwen2.5-Instruct & 14\,B
 & 74.0 & 74.0 & 74.0 & 75.0 & 73.0 \\

 & \textbf{Ensemble Model} & 55\,B\textsuperscript{**}
 & \textbf{78.5} & \textbf{78.4} & 77.1 & 83.7 & 72.9 \\

\bottomrule

\multicolumn{8}{@{}p{\textwidth}@{}}{\footnotesize
\textsuperscript{*}Parameter counts for GPT are not officially
disclosed by OpenAI; the reported values are unofficial estimates. \textsuperscript{**} 55\,B is the total number of parameters across the five
fine-tuned models used for ensemble voting.}
\end{tabular}
\end{table*}

\subsection{Agreement with Benchmark Labels}

\begin{figure*}[tb]
    \centering
    \includegraphics[trim={0cm 0.3cm 0cm 0.5cm}, clip,width=\textwidth]{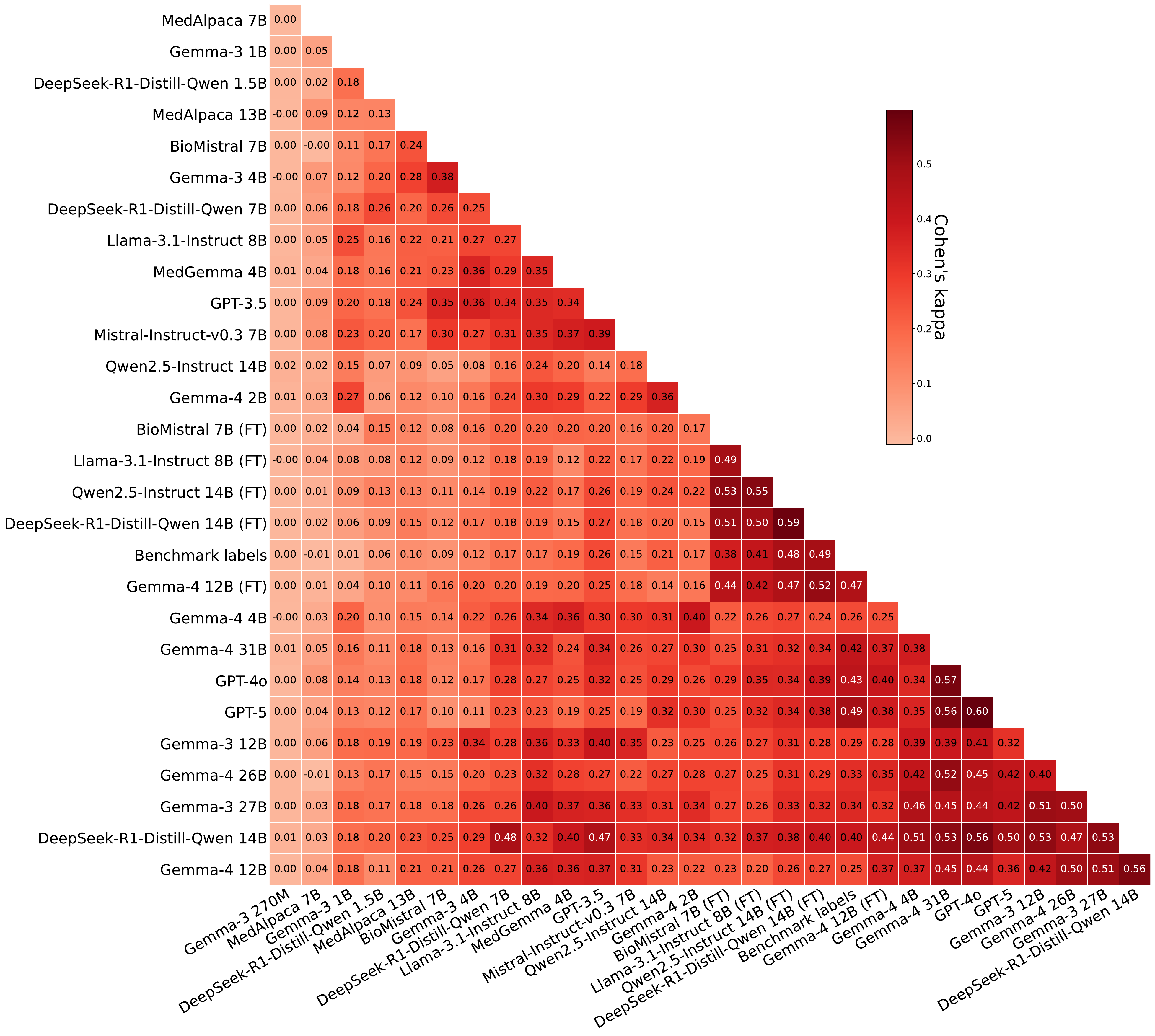}
    \caption{\small{Pairwise Cohen's $\kappa$ agreement between model predictions and the
benchmark reference labels, computed on the held-out benchmark test set
($n=1{,}050$). Higher values (darker) indicate stronger agreement. Values quantify agreement with the benchmark rather than
clinical correctness.}}\vspace{-0.15in}
    \label{fig:kappa_heatmap}
\end{figure*}

We further assessed inter-model agreement on ECGQuest using pairwise Cohen’s $\kappa$ between model predictions and the benchmark reference labels (Figure~\ref{fig:kappa_heatmap}). Because the questions and reference answers were generated and evaluated using language models, these values measure agreement with the benchmark or between models, not clinical correctness. Overall, agreement with the reference labels generally follows the accuracy ranking. Among models evaluated in zero-shot, the agreement is highest for GPT-5 ($\kappa=0.49$), GPT-4o (0.43), followed by Gemma 4-31\,B (0.42), and DeepSeek-R1-14\,B (0.40). In contrast, strongly biased models show almost no agreement despite accuracies near 50\%, including MedAlpaca-7\,B ($-0.01$) and Gemma-3-270\,M (0.00). Fine-tuning leads to the biggest improvements; the largest of these occured in the models that started as weakest: agreement rises from 0.09 to 0.38 for BioMistral-7\,B, from 0.17 to 0.41 for Llama-3.1-Instruct-8\,B, from 0.21 to 0.48 for Qwen2.5-Instruct-14\,B, from 0.25 to 0.47 for Gemma-4-12\,B, and from 0.40 to 0.49 for DeepSeek-R1-14\,B. All five fine-tuned models consequently rank among the strongest in the panel, with DeepSeek-R1-14\,B (FT) matching GPT-5 at 0.49 and Qwen2.5-Instruct-14\,B (FT) and Gemma-4-12\,B (FT) surpassing GPT-4o at 0.48 and 0.47. Inter-model agreement is likewise concentrated among the top performers. The two closed-models agree most strongly with each other ($\kappa=0.60$), and Gemma-4-31\,B agrees with GPT-4o and GPT-5 at 0.57 and 0.56, closely followed by DeepSeek-R1-14\,B at 0.56 and 0.50, indicating that the strongest models tend to make similar choices. The fine-tuned models form a separate cluster, agreeing more with one another ($\kappa=0.42$ to $0.59$) than with their base models. This shows that task-specific fine-tuning pushes different model types toward a shared, label-focused decision pattern.

Additional views of these results are provided in the supplement. Supplement~\ref{app:difficulty} reports a question-level difficulty and clarity profile, in which approximately $73\%$ of test items are answered correctly by at least half of evaluated models, with a small residual subset that almost none answers correctly. Supplement~\ref{app:category} reports category-level accuracy for the ten strongest models, where no single model leads across all four ECG knowledge categories and Fine-tuned models lead in most categories.

\subsection{Pareto Frontiers on Receiver Operating Characteristics and Precision Recall Curves}
\begin{figure*}[tb]
    \centering
    \includegraphics[width=1\textwidth]{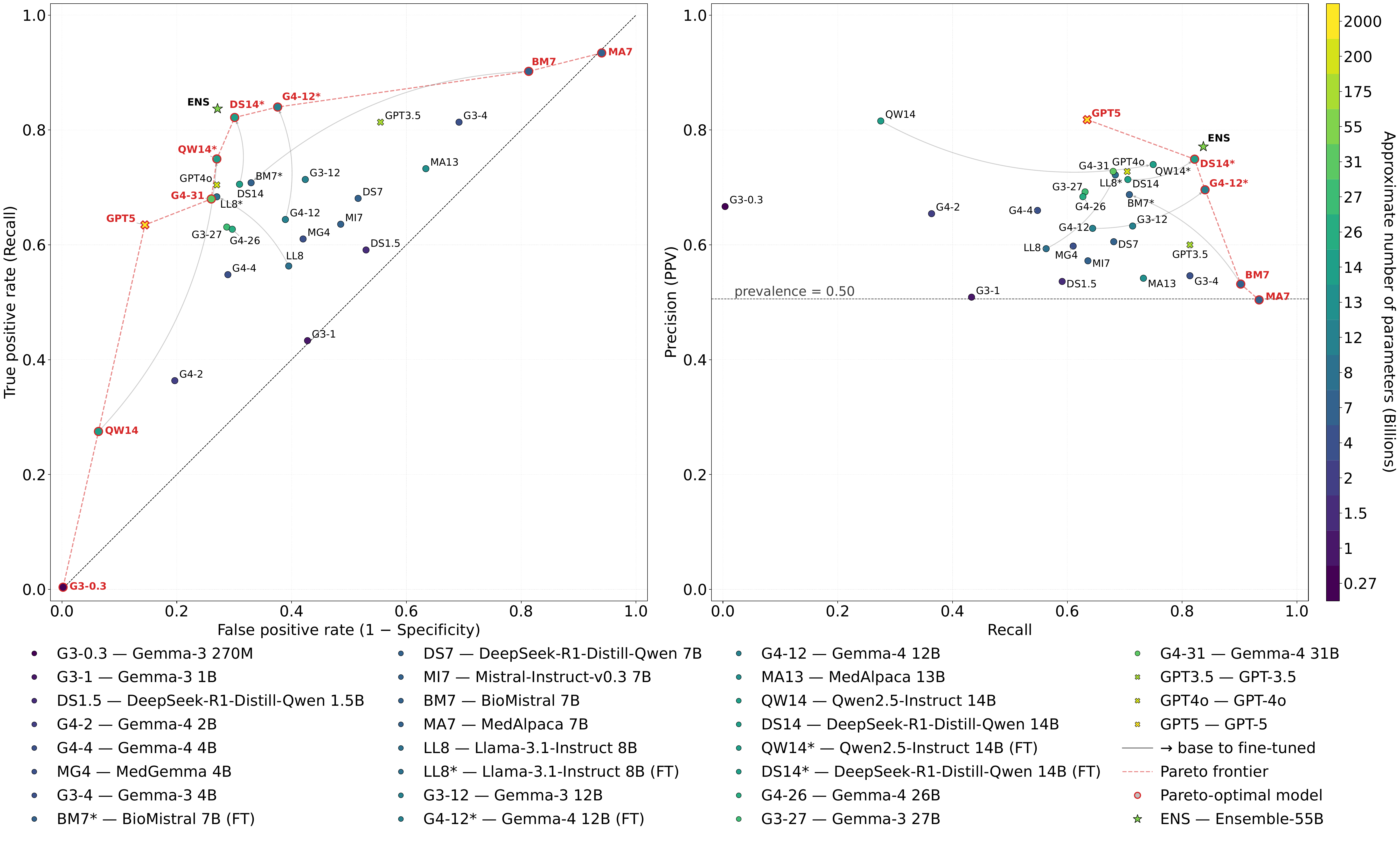}
    \caption{\small{Operating points of all evaluated models in ROC (left) and PR (right) spaces on ECGQuest. As each model make hard true/false decisions, it contributes a single operating point rather than a curve. Marker color encodes the approximate parameter count; circles denote open source models and crosses closed source models. Grey arrows connect each base model to its fine-tuned counterpart (FT), red dashed lines trace the Pareto frontier, and red rings mark Pareto-optimal models. The black dashed lines indicate chance performance.}}\vspace{-0.15in}
    \label{fig:operating_points}
\end{figure*}

Any statistical classification task, including language-based tasks, is multiobjective in nature, and its performance reflects a balance between how sensitive and specific the model is. However, most language models do not provide us with access to their internal decision thresholds, which would allow us to change their operating points through the choice of parameters. Nonetheless, we can visualize their multiobjective tradeoff. Figure~\ref{fig:operating_points} shows each model as a single point in Receiver Operating Characteristic (ROC) space (left) and precision-recall (PR) space (right). We have also shown the models on the \textit{Pareto frontier}, i.e., models for which no other model performs at least as well on both objectives and strictly better on at least one. Since the fine-tuning objective was to maximize the macro-F1 score (equivalently maximizing accuracy because of the balanced True/False questions), we can envision each fine-tuned model as moving toward the ideal classifier in ROC and PR spaces. In ROC space, this brings the models closer to the upper-left corner and increases Youden's $J$ statistic ($J=\mathrm{sensitivity}+\mathrm{specificity}-1$), increasing the diagonal distance of an operating point above the chance diagonal. In Figure~\ref{fig:operating_points}, the gray arrows connect each base model to its fine-tuned counterpart. We can see that $J$ increases in every case: from 0.211 to 0.480 for Qwen2.5-Instruct-14\,B, whose recall rises from 27.5 to 75.0 at essentially unchanged specificity; from 0.09 to 0.38 for BioMistral-7\,B, which moves in the opposite direction as its specificity climbs from 18.7 to 67.1; and from 0.40 to 0.52 for DeepSeek-R1-14\,B. The ensemble model outperforms individual models, with a false-positive rate of 0.27 and a recall of 0.837, giving $J$=0.57, compared with GPT-5's 0.49. 

\section{Discussion and Future Work}\label{sec:discussion}
The results demonstrate that ECG knowledge varies across models and is not consistent even in medical-specific language models, highlighting the need for dedicated ECG benchmarks and ECG-specific language models, since general medical performance does not always predict how well a model will perform on ECG-specific tasks. Within the same architecture, performance generally improves with model size, as observed in both Gemma families. Across families, however, size does not always correlate with performance; pretraining data and base-model quality matter at least as much. Medical-specific models are a clear example: medically fine-tuned models under-performed larger general-purpose models, and MedGemma-4\,B, the only competitive medical model, is outperformed by Gemma-4-4\,B at identical parameter size. This suggests that medical-specific tuning does not necessarily provide the domain-specific ECG knowledge required for this task.




On a balanced True/False dataset, accuracy provides an overall measure of performance but may obscure class-specific imbalance. For example, a model that always predicts one class achieves \(50\%\) accuracy but only \(33.3\%\) macro-F1 because its F1 score for the other class is zero.  We therefore report and compare macro-F1 alongside accuracy. A large gap between accuracy and macro-F1 indicates asymmetric performance across the True and False classes, although the direction of the bias must be determined from the class-specific results. Although several models exhibit a bias toward either True or False responses, GPT-5, GPT-4o, Llama-3.1-Instruct-8\,B, MedGemma-4\,B, and the larger Gemma-3 and Gemma-4 models have balanced performance across the two classes.


Fine-tuning SLMs closed the gap between open-source SLMs and commercial LLMs such as GPT-5 and GPT-4o, the two strongest models in the zero-shot evaluation. All five adapted models outperform the encoder baselines by a wide margin and match or exceed GPT-4o; while DeepSeek-R1-14\,B and the majority vote over the five models exceed GPT-5. That a 14\,B model matches a proprietary model of far larger scale indicates that the ECG
knowledge can be transferred into a compact model. Also, because of the training setup and balanced dataset we created, adaptation mitigated the initial model class biases, moving their operating point toward the Youden's $J$-statistics.

On the external exam-labeled sets (Supplement~\ref{app:external}), LoRA improved performance and corrected directional bias in BioMistral-7\,B and Qwen2.5-14\,B, but for balanced base models such as Llama-3.1-Instruct-8\,B or DeepSeek-R1-14\,B it mainly specialized the model toward the ECGQuest distribution, reducing accuracy. We read this as a generalization gap rather than a training failure. The external sets also warrant caution: MedMCQA and MedQA are long-standing public resources. Their contents have likely entered the training corpora of existing models; so strong zero-shot scores there may partly reflect data leakage rather than ECG understanding.

As a limitation of our work, although ECGQuest questions and labels were generated by GPT-4o from ECG reference materials and grounded in the provided source text, the labels have not undergone expert over-read and should therefore be considered weak labels rather than clinical ground truth. 
To facilitate future expert validation, each ECGQuest item is linked to a quote from its source.

Future work should expand ECGQuest with additional public resources, including clinical guidelines, educational materials, open-access articles, and abstracts from cardiology conference proceedings. Furthermore, items should be over-read by human experts to identify ambiguous items, correct clinically inaccurate labels, and establish an expert-validated benchmark. Because each ECGQuest item is linked to its source text, expert review can be conducted systematically while maintaining full traceability. ECGQuest can then be paired with models that operate directly on waveforms or images,
so that measurement-based models supply the quantitative findings while an ECG-specific
language model interprets them using electrophysiological and clinical knowledge, with
expert review closing the loop. Also, the data preparation, fine-tuning, and evaluation
pipeline is model-agnostic and can be reapplied as new models become available. Future studies can extend this framework to additional models and cardiology modalities, including echocardiography, cardiac magnetic resonance imaging (MRI), and cardiac computed tomography (CT), enabling specialized models that integrate complementary cardiovascular information.

\section{Conclusion}
We developed ECGQuest, a literature-grounded balanced True/False dataset derived from ECG references for evaluating and fine-tuning language models on conceptual ECG knowledge. Each item is source-linked and paired with a minimally modified opposite-label counterpart. The evaluation produced three main findings. First, ECG knowledge varies substantially across models and is not guaranteed by generic medical-specialized models. Second, within the same model family, performance generally improves with increasing model size. Third, parameter-efficient fine-tuning enables relatively small $7$--$14$\,B models to match or outperform substantially larger proprietary models on ECGQuest, though evaluation on external exam labeled ECG questions (Supplement~\ref{app:external}) shows that this benefit is largest for weak or class-biased base models. The ECGQuest Q\&A set and the fine-tuned models presented in this work are publicly available\footnote{\textcolor{blue}{Note: The dataset and developed models will be made public and their URL will be added to the paper upon manuscript acceptance.}}.




\bibliographystyle{iclr2026_conference}

\bibliography{refs}

\lstdefinestyle{ieeeprompt}{
  basicstyle=\ttfamily\footnotesize,
  breaklines=true,
  breakindent=0pt,
  columns=fullflexible,
  keepspaces=true,
  frame=single,
  framerule=0.4pt,
  framesep=5pt,
  xleftmargin=5.4pt,     
  xrightmargin=5.4pt,    
  aboveskip=2pt,
  belowskip=2pt,
}

\newpage

\clearpage
\appendix 
\appendix
\section*{Supplementary Materials}
\section{ECGQuest Question Composition and Level of Difficulty}\label{app:ecgquest}
We review ECGQuest in terms of dataset composition, question difficulty, and model performance. The analysis includes the distribution of question pairs across dataset splits and ECG knowledge categories, the percentage of cases in which each question is answered correctly across tested models, and the performance of the top-performing models within each knowledge category.

\textit{Dataset Composition:}\label{app:dataset}
GPT4o was queried to categorize ECGQuest under four categories listed in 
Table~\ref{tab:dataset_stats}. The table reports the number of ECGQuest question pairs in each category across the training, validation, and test splits.

\begin{table}[tbh]
\centering
\caption{Distribution of benchmark question pairs across dataset splits and question categories.}
\label{tab:dataset_stats}
\begin{tabular}{lrrrr}
\toprule
\textbf{Question Category} & \textbf{Train} & \textbf{Validation} & \textbf{Test} & \textbf{Total} \\
\midrule
ECG Diagnosis 
    & 5{,}300 & 703 & 732 & 6{,}735 \\
ECG Characteristics
    & 1{,}499 & 224 & 194 & 1{,}917 \\
ECG Acquisition Technologies
    & 1{,}045 & 74 & 81 & 1{,}200 \\
Basic Cardiac Physiology
    & 959 & 50 & 43 & 1{,}052 \\
\midrule
\textbf{All Types}
    & \textbf{8{,}803}
    & \textbf{1{,}051}
    & \textbf{1{,}050}
    & \textbf{10{,}904} \\
\bottomrule
\end{tabular}%
\end{table}

\begin{figure}[tbh]
\centering
    \includegraphics[width=0.6\columnwidth]{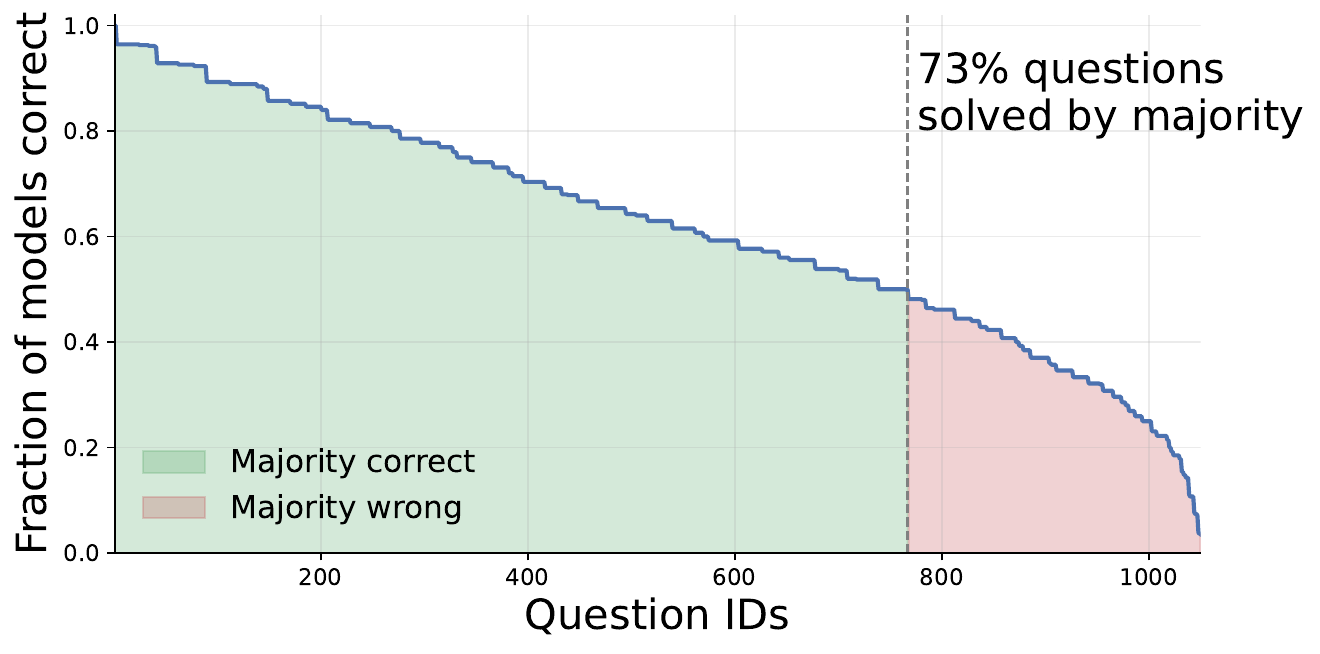}
    \caption{Question-level difficulty/clarity profile on the ECGQuest test set
    ($n=1{,}050$). Questions are ordered from easiest/clear to hardest/ambiguous according to the fraction of evaluated models that answered each question correctly. The green region represents questions answered correctly by at least half of the models, while the red region represents questions answered correctly by fewer than half of the models. The dashed vertical line indicates that approximately $73\%$ of the questions were answered correctly by at least half of the evaluated models.}
    \label{fig:difficulty_profile}
\end{figure}

\textit{Question-Level Difficulty Profile:}
\label{app:difficulty}
At the question level, some questions were easier or harder for the models to answer, as reflected in the degree of agreement across model responses. Without expert review, this can also be a reflection of more clear or more ambiguous questions. Figure~\ref{fig:difficulty_profile} shows the fraction of models that answered each benchmark question correctly, with questions ordered from easiest/clear to hardest/ambiguous. At least half of the evaluated models answered approximately $73\%$ of the 1{,}050 test questions correctly, whereas fewer than half answered the remaining 27\% correctly. The gradual decline across the ranked questions indicates substantial variation in question difficulty/clarity, ranging from items answered correctly by nearly all models to a small subset that almost none answered correctly.

\begin{figure}[tb]
    \centering
    \includegraphics[width=0.95\columnwidth]{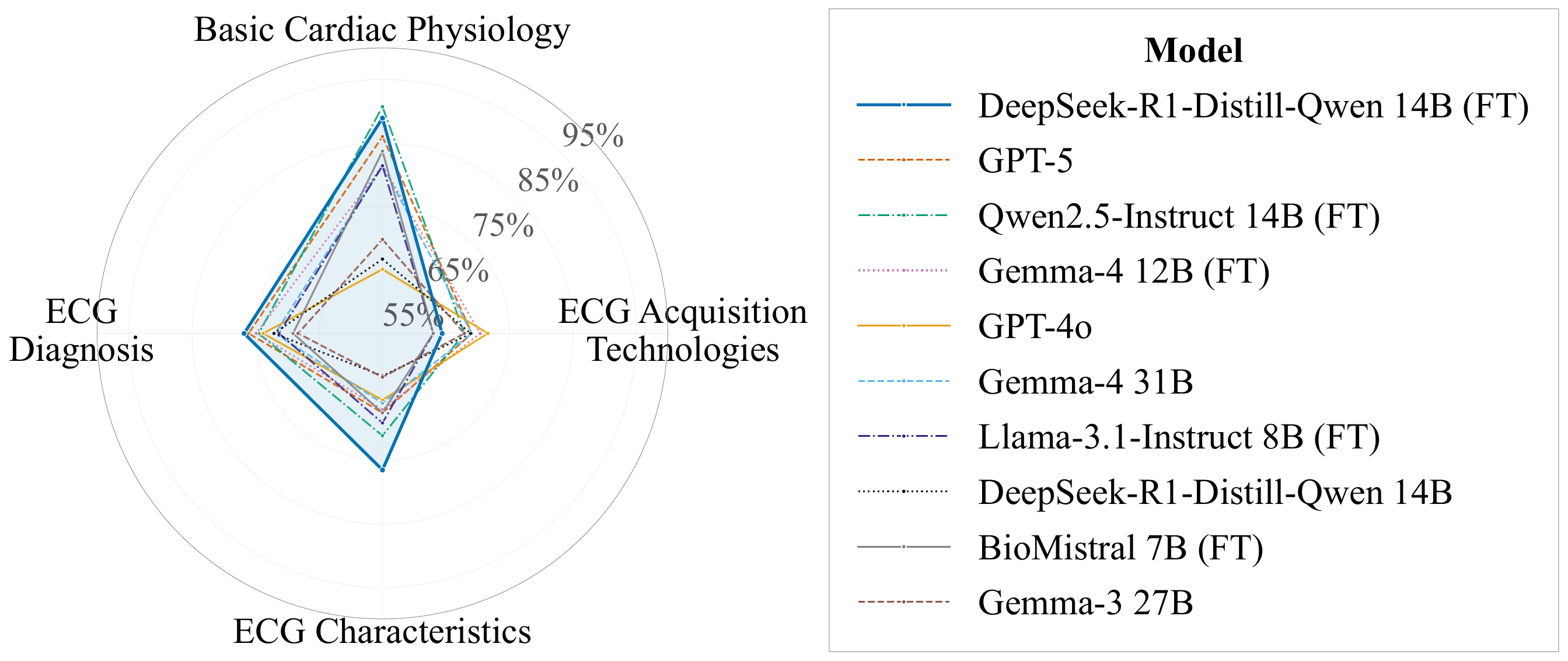}
    \caption{Category-level accuracy of the ten models with the highest overall performance across four ECG knowledge categories: ECG Diagnosis, ECG Acquisition Technologies, ECG Characteristics, and Basic Cardiac Physiology.}
    \label{fig:category_radar}
\end{figure}

\textit{Category-Level Performance:}
\label{app:category}
Figure~\ref{fig:category_radar} compares the category-level accuracy of the ten models with the highest overall performance. The results show that no single model achieved the best performance across all ECG knowledge categories. In \textit{ECG Diagnosis}, the fine-tuned DeepSeek-R1-Distill-Qwen 14\,B model achieved the highest accuracy at 76.8\%, followed by GPT-5 at 76.1\%. For \textit{ECG Acquisition Technologies}, GPT-4o ranked first with 71.6\%, while the fine-tuned Gemma-4 12\,B model followed closely with 70.4\%. In \textit{ECG Characteristics}, the fine-tuned DeepSeek-R1-Distill-Qwen 14\,B model achieved the highest accuracy of 76.5\%, followed by the fine-tuned Qwen2.5-Instruct 14\,B model at 71.1\%. The strongest overall category-level performance was observed in \textit{Basic Cardiac Physiology}, where the fine-tuned Qwen2.5-Instruct 14\,B model achieved the highest accuracy at 90.7\%, followed closely by the fine-tuned DeepSeek-R1-Distill-Qwen 14\,B model at 88.9\%. Overall, these results highlight complementary strengths across models and suggest that fine-tuning was particularly beneficial.

\section{External Exam-Labeled Evaluation}\label{app:external}
As supplementary analysis, in order to assess model generalization beyond ECGQuest, we evaluated all models on ECG-related questions from standard medical tests, as detailed below.

\textit{Dataset Preparation:}\label{app:external-prep}
ECG-related questions were extracted from multi-option Q\&A exams MedMCQA~\citep{pal2022medmcqa} and MedQA~\citep{jin2021medqa}. Identical ECG-topic filtering was applied to both datasets. For this, GPT-4o was used as a strict topic classifier to identify questions within the four ECG topics covered by the benchmark. This resulted in 1,054 questions from 187,005 in MedMCQA and 150 out of 12,723 in MedQA. Questions with negated or except-style format were excluded, as these could not be converted into a single unambiguous True/False statement (12 removed from MedMCQA and 3 from MedQA). The multi-option questions were next appended with the correct option text (from the exam key). Next, to achieve a balanced dataset, a false version of each question was created by appending the question with a randomly picked wrong response from the false options. For example, the question \textit{``P wave is absent in:''} with options \textit{A) Atrial fibrillation, B) Atrial asystole, C) Ventricular fibrillation, and D) Ventricular tachycardia}, with the correct answer A, was converted into the following two True and False question-answer pairs:
\begin{itemize}
    \item True version: \textit{P wave is absent in Atrial fibrillation.}
    \item False version: \textit{P wave is absent in Ventricular tachycardia.}
\end{itemize}
where the false answer was picked randomly between the three incorrect options. This procedure produced 2,084 MedMCQA and 294 MedQA True/False questions. As the labels are derived from the official examination key, these sets provide a measure of ECG knowledge independent of the question generation process.

\textit{Zero-Shot and Fine-Tuned Results:}
\label{app:external-results}
Table~\ref{tab:external_table} presents the results on ECG-related MedMCQA and MedQA Q\&A (Section~\ref{app:external-prep}). The ranking of models are largely similar to the ECGQuest ordering. GPT-5 leads on both (86.5\% on MedMCQA, 93.2\% on MedQA), followed by GPT-4o (83.7\%, 89.8\%) and then Gemma-4-31\,B (79.7\%, 83.0\%) and Gemma-4-26\,B (75.4\%, 81.0\%), which both outperform GPT-3.5 (70.4\%, 67.3\%). Both Gemma families also show a monotonic increase in accuracy with model size on both sets. Among the medical models MedGemma-4\,B is the only competitive medical model (66.8\%, 60.9\%), while MedAlpaca-7\,B, BioMistral-7\,B, and MedAlpaca-13\,B perform near chance on both sets (from $49.4$ to $52.3\%$).

On the external datasets, the effect of fine-tuning was highly dependent on the base model. For the low-performing biased BioMistral-7\,B, fine-tuning improved accuracy on both sets (from 51.3\% to 60.5\% on MedMCQA; from 50.7\% to 58.5\% on MedQA) and roughly doubled macro-F1 (from 36.7 to 60.3 and from 35.4 to 58.4) by correcting the True class bias. Qwen2.5-Instruct-14\,B, which had an opposite bias, improved on MedMCQA in both accuracy (67.3\% to 73.0\%) and macro-F1 (64.6 to 72.9), while remaining close to its starting point on MedQA (66.3\% to 64.9\%). Gemma-4-12\,B remained close to its base on MedMCQA (74.4\% to 72.6\%) and improved on MedQA (68.7\% to 72.8\%). For the already strong Llama-3.1-8\,B and DeepSeek-R1-14\,B models, fine-tuning reduced accuracy on both external sets (from 71.1\% to 63.2\% and from 70.8\% to 65.5\% on MedMCQA; from 66.7\% to 55.8\% and from 70.4\% to 62.6\% on MedQA), and the ensemble built from all five models shows the same pattern (67.2\% and 65.6\%). This can be related to the fact that MedMCQA and MedQA are long-standing public resources whose items have likely entered the pretraining corpora of the evaluated models, so strong zero-shot scores may partly reflect data leakage rather than ECG understanding.

\begin{table*}[t]
\centering
\caption{Zero-shot and fine-tuned performance on the external, exam-labeled
MedMCQA ($n{=}1{,}042$) and MedQA ($n{=}147$) True/False sets.
Acc: accuracy; F1: macro-averaged F1 score, P: precision, and R: recall;
Sen: sensitivity, Spec: specificity.
Across all models, the best value per metric column is in \textbf{bold} and the
runner-up is \underline{underlined}. The results can be compared and contrasted with the results in Table~\ref{tab:zeroshot} of the main text.
}
\label{tab:external_table}
\resizebox{\textwidth}{!}{%
\begin{tabular}{clrccccccccccc}
\toprule
 & & & \multicolumn{5}{c}{\textbf{MedMCQA}~\citep{pal2022medmcqa}} & \multicolumn{5}{c}{\textbf{MedQA}~\citep{jin2021medqa}} \\
\cmidrule(lr){4-8}\cmidrule(lr){9-13}
 & \textbf{Model} & \textbf{Params} & \textbf{Acc (\%)} & \textbf{F1} & \textbf{P} & \textbf{R}/\textbf{Sen} & \textbf{Spec}
 & \textbf{Acc (\%)} & \textbf{F1} & \textbf{P} & \textbf{R}/\textbf{Sen} & \textbf{Spec} \\
\midrule
\multirow{26}{*}{\rotatebox[origin=c]{90}{\textbf{Zero-shot learning}}}
 & \multicolumn{12}{l}{\textit{General purpose (closed source)}} \\
 & GPT-5 & 2\,T\textsuperscript{*} & \textbf{86.5} & \textbf{86.5} & \textbf{88.8} & 83.5 & \underline{89.4} & \textbf{93.2} & \textbf{93.2} & \textbf{97.0} & 89.1 & \textbf{97.3} \\
 & GPT-4o & 200\,B\textsuperscript{*} & \underline{83.7} & \underline{83.7} & 81.6 & 86.9 & 80.4 & \underline{89.8} & \underline{89.8} & 89.3 & \underline{90.5} & 89.1 \\
 & GPT-3.5 & 175\,B\textsuperscript{*} & 70.4 & 70.3 & 68.1 & 76.8 & 64.1 & 67.3 & 67.1 & 64.7 & 76.2 & 58.5 \\
\cmidrule(l){2-13}
 & \multicolumn{12}{l}{\textit{General-purpose (open source)}} \\
 & Mistral-Instruct-v0.3 & 7\,B & 62.7 & 61.4 & 59.3 & 81.0 & 44.3 & 55.4 & 53.6 & 53.9 & 75.5 & 35.4 \\
 & Llama-3.1-Instruct & 8\,B & 71.1 & 71.1 & 71.2 & 70.8 & 71.4 & 66.7 & 66.6 & 68.1 & 62.6 & 70.7 \\
 & Qwen2.5-Instruct & 14\,B & 67.3 & 64.6 & \underline{88.1} & 39.9 & \textbf{94.6} & 66.3 & 62.9 & \underline{91.4} & 36.1 & \underline{96.6} \\
 & DeepSeek-R1-Distill-Qwen & 1.5\,B & 51.8 & 51.7 & 51.6 & 57.4 & 46.3 & 46.9 & 46.8 & 46.9 & 46.9 & 46.9 \\
 & DeepSeek-R1-Distill-Qwen & 7\,B & 51.7 & 51.7 & 51.9 & 48.4 & 55.1 & 56.5 & 56.5 & 56.5 & 56.5 & 56.5 \\
 & DeepSeek-R1-Distill-Qwen & 14\,B & 70.8 & 70.8 & 69.3 & 74.9 & 66.8 & 70.4 & 70.2 & 70.9 & 70.4 & 70.4 \\
 & Gemma-3 & 270\,M & 52.0 & 51.8 & 52.4 & 44.7 & 59.3 & 49.0 & 48.9 & 48.9 & 44.2 & 53.7 \\
 & Gemma-3 & 1\,B & 54.1 & 52.6 & 53.0 & 72.0 & 36.3 & 50.7 & 50.6 & 50.6 & 55.8 & 45.6 \\
 & Gemma-3 & 4\,B & 61.8 & 61.8 & 62.1 & 60.5 & 63.1 & 53.1 & 53.0 & 52.8 & 57.1 & 49.0 \\
 & Gemma-3 & 12\,B & 67.3 & 66.8 & 64.0 & 78.9 & 55.7 & 66.3 & 66.3 & 66.2 & 66.7 & 66.0 \\
 & Gemma-3 & 27\,B & 70.4 & 69.8 & 65.7 & 85.4 & 55.5 & 69.0 & 68.9 & 66.9 & 75.5 & 62.6 \\
 & Gemma-4 & 2\,B & 61.5 & 60.9 & 65.5 & 48.8 & 74.3 & 59.2 & 56.3 & 69.0 & 33.3 & 85.0 \\
 & Gemma-4 & 4\,B & 68.7 & 67.9 & 77.3 & 53.0 & 84.5 & 66.7 & 65.9 & 73.8 & 51.7 & 81.6 \\
 & Gemma-4 & 12\,B & 74.4 & 74.3 & 77.5 & 68.7 & 80.0 & 68.7 & 67.6 & 79.6 & 50.3 & 87.1 \\
 & Gemma-4 & 26\,B & 75.4 & 75.4 & 73.5 & 79.5 & 71.4 & 81.0 & 80.9 & 78.6 & 85.0 & 76.9 \\
 & Gemma-4 & 31\,B & 79.7 & 79.6 & 78.0 & 82.5 & 76.8 & 83.0 & 82.9 & 87.0 & 77.6 & 88.4 \\
\cmidrule(l){2-13}
 & \multicolumn{12}{l}{\textit{Medical-specialized (open source)}} \\
 & MedGemma & 4\,B & 66.8 & 66.2 & 63.3 & 79.8 & 53.7 & 60.9 & 58.3 & 57.3 & 85.7 & 36.1 \\
 & MedAlpaca & 7\,B & 49.4 & 33.0 & 49.7 & \underline{98.7} & 0.1 & 49.7 & 33.2 & 49.8 & \textbf{99.3} & 0.1 \\
 & BioMistral & 7\,B & 51.3 & 36.7 & 50.7 & \textbf{99.4} & 3.3 & 50.7 & 35.4 & 50.3 & \textbf{99.3} & 2.0 \\
 & MedAlpaca & 13\,B & 52.3 & 44.7 & 51.3 & 89.4 & 15.2 & 48.3 & 42.1 & 49.0 & 81.0 & 15.6 \\
\midrule
 & \multicolumn{12}{l}{\textit{Encoder baselines}} \\
 & BERT & 110\,M & 49.4 & 48.3 & 49.6 & 64.1 & 34.7 & 49.3 & 47.2 & 49.5 & 69.4 & 29.3 \\
 & BiomedBERT & 110\,M & 50.6 & 50.6 & 50.6 & 50.9 & 50.3 & 48.0 & 44.9 & 48.6 & 71.4 & 24.5 \\
\cmidrule(l){2-13}
\multirow{7}{*}{\rotatebox[origin=c]{90}{\textbf{Fine-tuned}}}
 & \multicolumn{12}{l}{\textit{Decoder LLMs}} \\
 & BioMistral & 7\,B & 60.5 & 60.3 & 59.2 & 67.6 & 53.4 & 58.5 & 58.4 & 59.5 & 53.1 & 63.9 \\
 & Llama-3.1-Instruct & 8\,B & 63.2 & 63.2 & 64.6 & 58.7 & 67.8 & 55.8 & 53.1 & 61.0 & 32.0 & 79.6 \\
 & Gemma-4 & 12\,B & 72.6 & 72.3 & 69.2 & 81.2 & 63.9 & 72.8 & 72.8 & 71.6 & 75.5 & 70.1 \\
 & DeepSeek-R1-Distill-Qwen & 14\,B & 65.5 & 65.4 & 63.9 & 71.4 & 59.7 & 62.6 & 62.6 & 62.6 & 62.6 & 62.6 \\
 & Qwen2.5-Instruct & 14\,B & 73.0 & 72.9 & 76.0 & 67.2 & 78.8 & 64.9 & 64.9 & 64.9 & 65.0 & 64.8 \\
 & \textbf{Ensemble Model} & 55\,B\textsuperscript{**} & 67.2 & 67.1 & 65.5 & 72.6 & 61.8 & 65.6 & 65.5 & 68.0 & 59.2 & 72.1 \\
\bottomrule
\multicolumn{13}{@{}p{1.3\textwidth}@{}}{\footnotesize
\footnotesize{*} Parameter counts for the GPT models are not officially disclosed by OpenAI; the reported values are unofficial estimates. \footnotesize{**} 55\,B is the total number of parameters across the five fine-tuned models used for ensemble voting.}
\end{tabular}
}
\end{table*}

\section{Question-Generation, Validation, and Label-Balancing Prompts}\label{app:prompts}

For reproducibility, we provide the three prompts used to construct the Q\&A sets described in Section~\ref{sec:data_cons}. Generation used two modes depending on page content: a text prompt (Prompt~\ref{prompt:text_prompt}) for pages processed from extracted text, and a vision prompt (Prompt~\ref{prompt:vision_prompt}) for pages processed from a rendered image. A separate label-balancing prompt (Prompt~\ref{prompt:negation_prompt}) generated the paired opposite label counterpart for each item (Section~\ref{sec:tf_balance}).

 

\captionof{prompt}{Text-mode generation prompt. GPT-4o reads only the extracted page text.}
\begin{lstlisting}[style=ieeeprompt]
You are an expert medical exam question writer specializing in ECG interpretation. Generate clinical TRUE/FALSE questions for medical students, biomedical engineers, cardiologists, and electrophysiologists.

Generate questions only about the following allowed topics:

1. ECG waveform features, including P, QRS, ST, T, and U waves; QT, PR, and RR intervals; morphology; and axis deviation.

2. Clinical interpretation and diagnostic significance of ECG patterns, including ischemia, infarction, arrhythmias, conduction blocks, hypertrophy, pericarditis, electrolyte effects on the ECG, and criteria for distinguishing normal from abnormal findings.

3. The physiological basis of ECG findings, including the conduction system, depolarization and repolarization, action potentials, automaticity, and refractory periods.

4. ECG device technical specifications, including leads, electrode placement, gain, filters, paper speed, calibration, and artifacts.

STRICT RULES:

* Never hallucinate. Every question must come directly from the page text.
* Never use general medical knowledge. Use only facts stated in the text.
* `source_text` must contain the exact sentence or sentences, limited to one to three sentences, copied VERBATIM from the text.
* The question must be a DIRECT TRANSFORMATION of `source_text`. For a TRUE question, every word, number, and fact must be traceable to `source_text`. For a FALSE question, only the single altered detail may differ. Do not introduce any other outside knowledge, inference, or generalization.
* For a TRUE question, restate the key fact from `source_text`.
* For a FALSE question, change ONE detail, such as a number, threshold, or direction, and keep the rest from `source_text`.
* Discard any question if it contains content from outside `source_text`, except for the single detail intentionally changed in a FALSE question.
* Questions must be clinically meaningful and self-contained.
* Never reference images, figures, strips, diagrams, or tables.
* Balance the number of True and False answers.
* Return `[]` if the page contains no ECG-relevant content.

EXAMPLE:

source_text: "The normal PR interval ranges from 120 to 200 ms."

TRUE: "The normal PR interval ranges from 120 to 200 ms."

FALSE: "The normal PR interval ranges from 120 to 400 ms."

WRONG (outside knowledge): "A prolonged PR interval indicates first-degree AV block."

OUTPUT:

Return a Python list of dictionaries without Markdown:

[{"page":"page N", "question":"<transformation>", "label":"True", "source_text":"<verbatim sentence(s)>"}]

SELF-CHECK:

Discard the question if any of the following checks fail:

* For a TRUE question, `source_text` must support every word, number, and fact.
* For a FALSE question, all content must be supported by `source_text`, except for the single intentionally altered detail.
* `source_text` is copied verbatim.
* The label can be determined correctly from `source_text` alone.
* The question is self-contained.

Extract up to 30 questions.

[Extracted page text appended here.]
\end{lstlisting}
\label{prompt:text_prompt}


\captionof{prompt}{Vision-mode generation prompt. GPT-4o reads a rendered page image and uses only the text written on it.}
\begin{lstlisting}[style=ieeeprompt]
You are an expert medical exam question writer specializing in ECG interpretation. You are analyzing a PAGE IMAGE from an ECG textbook, manual, or clinical guide. Extract clinical knowledge as TRUE/FALSE exam questions.
 
Generate questions only about the following allowed topics:

1. ECG waveform features, including P, QRS, ST, T, and U waves; QT, PR, and RR intervals; morphology; and axis deviation.

2. Clinical interpretation and diagnostic significance of ECG patterns, including ischemia, infarction, arrhythmias, conduction blocks, hypertrophy, pericarditis, electrolyte effects on the ECG, and criteria for distinguishing normal from abnormal findings.

3. The physiological basis of ECG findings, including the conduction system, depolarization and repolarization, action potentials, automaticity, and refractory periods.

4. ECG device technical specifications, including leads, electrode placement, gain, filters, paper speed, calibration, and artifacts.
 
CRITICAL IMAGE RULES:

* Extract only clinical knowledge EXPLICITLY WRITTEN in the image, including captions, labels, tables, and annotations. Do NOT visually interpret waveforms or infer information from their shapes.
* Never use words or phrases that reference the image, including shown, figure, fig., strip, diagram, image, picture, above, below, indicated, arrow, highlighted, annotated, this ECG, this tracing, this rhythm, this waveform, this lead, this recording, as seen, or as depicted.
* BAD example: "In the rhythm strip shown, the RR interval is regular." GOOD example: "Regular RR intervals are a defining feature of normal sinus rhythm." Use this question only if the statement is explicitly written in a caption, label, table, or annotation.
  
STRICT RULES:

* Never hallucinate. Use only text written in the image. Do not use general medical knowledge.
* The question must be a DIRECT TRANSFORMATION of source_text. For a TRUE question, every word, number, and fact must be traceable to source_text. For a FALSE question, only the single altered detail may differ. Do not introduce any other outside knowledge, visual inference, or generalization.
* For a TRUE question, restate the written fact from source_text. For a FALSE question, change ONE detail, such as a number, threshold, or direction, and keep the rest from source_text.
* source_text must contain the exact text visible in the image, such as a caption, table cell, label, or annotation.
* Discard any question if it contains content from outside source_text, except for the single detail intentionally changed in a FALSE question.

OUTPUT:

Return a Python list of dictionaries without Markdown:
  [{"page":"page N", "question":"<transformation; no image reference>", "label":"True", "source_text":"<exact text visible in image>"}]
 
SELF-CHECK:

Discard the question if any of the following checks fail: * Source_text is written in the image. 
* For a TRUE question, source_text supports every word, number, and fact.
* The question contains no outside facts, visual inference, or generalization. 
* The label can be determined correctly from source_text alone.
* The question is self-contained and does not reference the image.

Extract up to 30 questions. Return [] if nothing relevant.
\end{lstlisting}
\label{prompt:vision_prompt}

During verification, each candidate item is checked against the generation constraints and discarded if it fails, with the reason recorded. The following text box shows two representative rejections: one whose cited span is not grounded in the page, and one whose question refers to a figure.

\noindent\textbf{Representative items removed during verification.}
\begin{lstlisting}[style=ieeeprompt]
% Rejection 1 -- reason: source_text_not_grounded

Question: "If a patient with atrial tachycardia is stable,
synchronised electrical cardioversion is recommended."

Label: False

source_text: "If the patient is stable, use vagal manoeuvres
and adenosine."

Reason: The statement is not supported by the cited text,
which recommends vagal manoeuvres and adenosine
rather than cardioversion. The question is discarded.

% Rejection 2 -- reason: image_reference_in_question

Question: "The QRS rhythm is irregular in Figure 4.7."

Label: False

source_text: "QRS rhythm: regular."

Reason: The question refers to Figure 4.7 and cannot be
answered without viewing the figure. The question
is discarded.

\end{lstlisting}

\label{prompt:rejected}


\captionof{prompt}{Label-balancing prompt. Given a validated item, GPT-4o produces the opposite-label counterpart from the same \emph{source\_text}.}
\begin{lstlisting}[style=ieeeprompt]
You are a medical exam question editor specializing in ECG interpretation.

TASK (original label = True -> produce a FALSE counterpart):

* Make minimal changes (such as a number, threshold, direction, or condition) or by paraphrasing, to make the statement clinically false.
* Do not introduce any concept that is not present in source_text.

TASK (label = False -> produce the TRUE counterpart):

* Restate the key fact from source_text as a correct true statement.
* Derive the statement only from source_text.


STRICT RULES:
  * Keep source_text IDENTICAL to the input. Do not change a single word.
  * Derive the counterpart ONLY from source_text. Do not introduce any outside fact or concept.
  * The counterpart must be a self-contained clinical statement.
  * Never reference images, figures, strips, or diagrams.

INPUT:
  original_question : <question>
  original_label    : <True|False>
  source_text       : <source_text>

OUTPUT (Python dict, no markdown):
  {"question":"<counterpart>", "label":"<target>",
   "source_text":"<identical to input>"}

SELF-CHECK:
Regenerate the counterpart if any of the following fail; if it still
fails after retries, discard it:
* source_text is identical to the input.
* The counterpart differs from the original question.
* The label is the correct inverse of the original label.
* For a FALSE counterpart, the statement is made clinically false either by changing one detail or by paraphrasing, while source_text is unchanged and no outside concept is introduced.
* For a TRUE counterpart, the statement is a faithful, correct
  restatement of source_text (no new or unsupported claim).
* The label can be determined correctly from source_text alone.
* The counterpart introduces no concept outside source_text.
* The counterpart is a self-contained clinical statement.
* The counterpart does not reference any image, figure, strip, or diagram.
\end{lstlisting}

\label{prompt:negation_prompt}

\end{document}